\documentclass[conference]{IEEEtran}
\IEEEoverridecommandlockouts
\usepackage{cite}
\usepackage{amsmath,amssymb,amsfonts}
\usepackage{algorithmic}
\usepackage{graphicx}
\usepackage{textcomp}
\usepackage{xcolor}
\def\BibTeX{{\rm B\kern-.05em{\sc i\kern-.025em b}\kern-.08em
    T\kern-.1667em\lower.7ex\hbox{E}\kern-.125emX}}
\begin{document}


\title{Comparative study of subset selection methods for rapid prototyping of 3D object detection algorithms
\thanks{The work presented in this paper was supported by the AGH University of Krakow project no. 16.16.120.773.}
}


\author{\IEEEauthorblockN{Konrad Lis}
\IEEEauthorblockA{\textit{Embedded Vision Systems Group,} \\
\textit{Department of Automatic Control and Robotics} \\
\textit{AGH University of Krakow}, Poland \\
kolis@agh.edu.pl}
\and
\IEEEauthorblockN{Tomasz Kryjak, \textit{Senior Member IEEE} }
\IEEEauthorblockA{\textit{Embedded Vision Systems Group,} \\
\textit{Department of Automatic Control and Robotics} \\
\textit{AGH University of Krakow}, Poland \\
tomasz.kryjak@agh.edu.pl}
}

\maketitle

\begin{abstract}

Object detection in 3D is a~crucial aspect in the context of autonomous vehicles and drones. 
However, prototyping detection algorithms is time-consuming and costly in terms of energy and environmental impact.
To address these challenges, one can check the effectiveness of different models by training on a~subset of the original training set. 
In this paper, we present a~comparison of three algorithms for selecting such a~subset --  \textit{random sampling},  \textit{random per class sampling}, and our proposed \textit{MONSPeC} (Maximum Object Number Sampling per Class). 
We provide empirical evidence for the superior effectiveness of random per class sampling and MONSPeC over basic random sampling. 
By replacing random sampling with one of the more efficient algorithms, the results obtained on the subset are more likely to transfer to the results on the entire dataset.
The code is available at: \textit{https://github.com/vision-agh/monspec}.
\end{abstract}

\begin{IEEEkeywords}
LiDAR, point cloud, object detection, PointPillars, CenterPoint, subset selection, MONSPeC, random per class sampling
\end{IEEEkeywords}

\section{Introduction}
\label{sec:introduction}

Advanced Driver Assistance Systems (ADAS), Autonomous Vehicles (AVs), and  Unmanned Aerial Vehicles (UAVs) rely on object detection for obstacle avoidance, traffic sign recognition, and object tracking. 
Vision stream processing is the most common approach, but LiDAR is sometimes preferred due to its resilience to lighting conditions and accurate 3D mapping. 
LiDAR is used in autonomous vehicles like Waymo and Mercedes S-Class. 
Working with LiDAR's 3D point clouds differs from vision systems due to the data format, which uses polar coordinates and includes reflection intensity information.

Object detection systems for autonomous vehicles commonly use three datasets: KITTI, Waymo Open Dataset, and NuScenes. 
The KITTI Vision Benchmark Suite (2012) \cite{Kitti} is the most widely used dataset, containing a~training set of $7481$ images, along with the corresponding point clouds and annotated objects. 
KITTI also maintains a~ranking of object detection methods.

The Waymo Open Dataset (2019) \cite{waymo} contains $1950$ sequences, corresponding to $200000$ frames, but only $1200$ sequences are annotated. 
However, it includes $12.6$ million objects. 
Waymo holds challenges in several topics each year, including 3D object detection and motion prediction.

NuScenes \cite{nuscenes} includes $1000$ sequences, which comprise approximately $1.4$ million images, $390$ thousand LiDAR scans, and $1.4$ million annotated objects.
It also maintains a~ranking of object detection methods.
From 390k LiDAR scans only 40k are annotated --
28310 are used for training, 6019 for validation and 6008 for testing.
Object detection methods on nuScenes are evaluated using standard mAP (mean Average Precision) metric and a metric called NDS (nuScenes detection score).
It includes mAP and several error measures, e.g. orientation error or scale error. 


There are two main approaches to object detection in point clouds: classical methods and those based on deep convolutional neural networks (DCNN). 
While DCNN-based approaches often achieve state-of-the-art results, they typically come with high computational and memory requirements.


Improving 3D LiDAR object detection involves academic and industrial research to enhance accuracy and real-time performance. 
This includes developing new architectures and optimizing hyperparameters using techniques like Neural Architecture Search and Hyperparameter Optimization, which require multiple learning processes.
Therefore, prototyping detection algorithms, especially with large datasets and/or complex computational tasks, requires a~lot of computation time. 
This leads to high energy consumption, costs, and environmental impact.
One solution to this problem is to work on a~subset of the training set instead of the entire set. This allows for computation time savings, which translates into reducing the aforementioned costs. 
However, careful selection of the subset is crucial to ensure comparable results to those obtained on the entire training set. 
This issue is the main focus of this study.

The main contributions of our work are:
\begin{itemize}
    \item we propose a simple deterministic algorithm for subset selection -- MONSPeC and compare it with other two algorithms for selecting a~subset of the training set,
    \item statistical analysis of the solutions with respect to the amount of training data in the subsets and deviation from the expected distribution of the number of objects in each class,
    \item presentation of empirical evidence for the superior effectiveness of MONSPeC and an other algorithm compared to completely random subset selection.
\end{itemize}
Experiments were conducted on the nuScenes dataset.

The reminder of this paper is organised as follows.
In Section \ref{sec:related_work} we discuss two issues related to our work: DCNN approaches to object detection in LiDAR data and subset selection methods.
Next, in Section \ref{sec:method} we elaborate on the motivation of our work, clarify the objective and present three subset selection algorithms.
The results obtained are summarised in Section \ref{sec:experiments}.
The paper ends with a~short summary with conclusions and discussion of possible future work.

\section{Related work}
\label{sec:related_work}

\subsection{3D object detection based on LiDAR data}
\label{ssec:3d_object_detection}

LiDAR-based 3D detectors can be categorized into point-based, voxel-based, and hybrid methods based on the representations of point clouds. 
Point-based approaches handle the point cloud in its original form, without any structure imposed. 
To process the point cloud, these methods first subsample the data and then employ DNNs based on PointNet++ \cite{pointnet++}. 
Point-RCNN \cite{pointrcnn} is an example of such an approach. 
Voxel-based methods involve voxelizing point clouds and processing the resulting tensor of voxels using 2D/3D DCNNs, as seen in methods such as PointPillars \cite{pointpillars} and VoxelNet \cite{voxelnet}.
Hybrid LiDAR-based 3D object detection methods combine elements from both voxel-based and point-based approaches. 
An example of such a~method is PV-RCNN \cite{pvrcnn}.

In this work, we perform experiments with two 3D object detection algorithms -- PointPillars and CenterPoint \cite{centerpoint} (version based on pillars).

\subsubsection{The PointPillars}
\label{sssec:pointpillars}

The PointPillars network \cite{pointpillars} is a~voxel-based 3D object detection method that employs a~pseudo-Bird's Eye View (BEV) map to replace the 3D convolutions with 2D convolutions, enabling end-to-end learning with less computational cost.
The point cloud is divided into a~set of ``pillars'', which are 3D cells in the XY plane. 
The first component of the network, the Pillar Feature Net (PFN), converts the point cloud into a~sparse ``pseudo-image''.
The second component is the Backbone, which is a~2D DCNN that processes the pseudo-image and extracts high-level features. 
The Backbone is composed of two subnets: the ``top-down'' subnet, which progressively reduces the dimension of the pseudo-image, and a~``bottom-up'' subnet that upsamples the intermediate feature maps and combines them into the final output map. 
Finally, the anchor-based Detection Head (Single-Shot Detector (SSD) network) performs object detection and regression of the 3D cuboids surrounding the objects on a~2D grid. 
The Non-Maximum Suppression (NMS) algorithm is then used to merge overlapping objects after inference.

\subsubsection{The CenterPoint}
\label{sssec:center_point}

CenterPoint \cite{centerpoint} is a~3D object detection method that uses either a~voxel-based or a~pillar-based representation. 
The input to CenterPoint is a~point cloud from a~LiDAR sensor.
The voxel-based and pillar-based versions of CenterPoint differ in type of backbone used.
The former uses voxel encoder and backbone from SECOND \cite{second}, the latter pillar encoder and backbone from PointPillars.
The output of the backbone is fed into the 1st stage detection head.
It consists of several regression maps -- one represents object centres as 2D gaussians on a~heatmap, others encode information such as sub-voxel location refinement, height above ground, 3D bounding box size and yaw rotation angle.
For each detected object, the second stage extracts 5 feature vectors from the backbone feature map, which correspond to the predicted object centre and centres of each of 4 sides of the bounding box.
The extracted feature vectors are concatenated and fed into an efficient MLP (MultiLayer Perceptron) which outputs a classification score and refines the bounding box.

\subsection{Selecting dataset subset}
\label{ssec:selecting_subset}

Machine learning algorithms typically require as much training data as possible to achieve the best performance. 
However, selecting a~subset of the training set and training algorithms on it also has its applications. 
One of them is Active Learning (AL), where unlabeled samples are gradually selected for labeling, aiming to maximise the improvement in model performance. 
In the object detection task on images, there are many approaches to AL, such as \cite{al_det_ensemble} and \cite{al_cls}.
They are based on ensemble of models, collecting their results and adding a~sample to the dataset if there is a~significant difference between the results of the models.
As for processing point clouds, existing AL approaches, \cite{lidar_al_2} do not focus directly on object detection tasks, but on semantic segmentation of point clouds. 
Other tasks, where selecting a~subset of the training set is useful, include Neural Architecture Search (NAS) or Hyperparameter Optimisation (HO). 
NAS involves automatically selecting the best network architecture in terms of performance, subject to constraints usually imposed on computational complexity or related factors such as speed. 
HO is a~broader term that also includes NAS. 
While NAS focuses on selecting hyperparameters directly related to network architecture, such as the number of layers or channels, HO focuses on selecting hyperparameters related to the entire training process (learning rate, weight decay, batch size and number of epochs). 
In these two tasks, working only on a~subset of the training set can significantly speed up the search for the best model.

In the context of point cloud processing, there are few works related to NAS or HO -- to our knowledge, the issue of NAS has only been addressed a~few times, e.g. \cite{lidar_nas1}.
However, the authors do not mention anything about working on a~subset of the training set. 
In the context of image processing, there are many more works.
Most of them focus on the classification task, and a~few of them use an approach with working on a~subset of the training set: \cite{nas_cls_proxy1}.
The majority of algorithms for obtaining such a~subset, like in AL, are based on detecting significant differences between the results of the ensemble of models.
Work focusing on the object detection task is NAS-FCOS \cite{nas_det_fcos}, the authors describe an efficient way to optimise the architecture for object detection. 
They use a~``carefully chosen subset of MS-COCO'', but the selection algorithm is briefly described.
Other papers, such as \cite{subsel_automata}, specifically address dataset subset selection but are tailored for image classification datasets and are not suitable for selecting samples in object detection.

\section{Proposed experimental method}
\label{sec:method}

\subsection{Motivation and objective}
\subsubsection{Motivation}
\label{sssec:motivation}

One of the tasks we are involved in is the search for fast and efficient algorithms for 3D object detection in LiDAR sensor point clouds. 
Training the CenterPoint-Pillar network on the entire nuScenes took about 20h (using two Nvidia RTX 3080 GPUs).
In contrast, using an~oversampled dataset as in \cite{cbgs}, the training took approximately 36h.
Prototyping an algorithm by repeatedly changing the architecture manually or searching automatically with NAS methods, such a~validation time for a~single architecture is far too long.
Longer computing times are associated with higher energy costs and higher environmental costs.
Therefore, we started to look for a~faster and less expensive way to compare different versions of the architectures.
A fairly obvious direction is to train models on a~subset of the original training set.

\subsubsection{Objective}
\label{sssec:objective}
The perfect subset of the training set for validating object detection architectures should meet two criteria:
\begin{enumerate}
    \item Be as small as possible
    \item If a~model A~trained on a~subset performs better than model B~trained on a~subset, then model A~trained on the whole set should also perform better.
\end{enumerate}

It is challenging, if not impossible, to provide condition 2 in the general case.
However, intuitively, the condition should be approximately fulfilled if the data distribution does not deviate too much from that of the entire dataset and there is as much training data (objects) as possible for each class in a~given subset.
For small subsets, the condition of maximising the number of objects is particularly important for sparse classes so as to reduce overfitting.
On the other hand, intuitively, the higher the detection efficiency for several architectures learned on a~subset relative to their detection efficiency obtained when training on the whole dataset, the better the subset reflects the training dataset and the higher the probability that condition 2 will be met.
We therefore propose the following procedure for creating a~subset:
\begin{enumerate}
    \item Assume a~subset size $N$ and select $N$ samples so as to maximise the number of objects for each class while not deviating too much from the distribution of the data across the dataset.
    Section \ref{ssec:algorithms_choosing_subset} presents 3 algorithms to implement this step, each presenting a~different trade-off between the number of objects and the distance from the original data distribution.
    \item Validate a~given subset by training several architectures on it and comparing their effectiveness with models trained on the whole set.
\end{enumerate}

\subsection{Subset selection algorithms}
\label{ssec:algorithms_choosing_subset}



Let us introduce the notations:
\begin{itemize}
    \item $N$ -- the number of samples in a~subset,
    \item $n$ -- the total number of objects in the subset,
    \item $n_k$ -- the number of objects of class $k$~in the subset,
    \item $p^{subset}_{k}$ -- the frequency of occurrence of objects of class $k$ in a~subset,
    \item $D$ -- the number of samples in the entire dataset,
    \item $d$ -- the total number of objects in the entire dataset,
    \item $d_k$ -- the number of objects of class $k$ in the entire dataset,
    \item $p^{dataset}_{k}$ -- the frequency of occurrence of objects of class $k$~in the entire dataset,
    \item $C$ -- the number of classes.
\end{itemize}

Certainly, the following equations apply: $n = \sum_ {k=1}^{k=C} n_k$, $p^{subset}_{k} = \frac{n_k}{n}$ and $\sum_ {k=1}^{k=C} p^{subset}_{k} = 1$.
Similarly: $d = \sum_ {k=1}^{k=C} d_k$, $p^{dataset}_{k} = \frac{d_k}{d}$
 and $\sum_ {k=1}^{k=C} p^{dataset}_{k} = 1$.


The most basic algorithm for selecting a~subset of the training set is a~completely random selection of samples -- \textit{random sampling}.
The expected value of the number of objects of class $k$ in a~subset is $\bar{n_k} = \frac{N}{D} \times d_k$, and the expected value of the total number of objects is $\bar{n} = \frac{N}{D} \times d$.
Thus, the expected ratio of the number of objects of a~class to the total number of objects is: $\frac{n_k}{n} = \frac{N}{D} \times d_k \times \frac{D}{N} \times \frac{1}{d} = \frac{d_k}{d}$.
Therefore, with high probability, the distribution of $p^{subset}_{k}$ with this type of sampling will be close to the distribution of $p^{dataset}_{k}$.


Another algorithm we are considering is the random selection of $\frac{N}{C}$ samples for each class separately -- \textit{random per class sampling}.
It is inspired by \textit{DS Sampling} from \cite{cbgs}, where it was used to balance the distribution of classes in the training set.
The steps of the algorithm are as follows:
\begin{enumerate}
    \item For each class, create a~set $idx_k$ which includes the indexes of training samples containing at least one object of that class.
    \item From each set $idx_k$, randomly select $\frac{N}{C}$ samples.
\end{enumerate}
It is worth noting that duplicate samples may appear in the resulting subset, as each sample may belong to several $idx_k$ sets.
Intuitively, the distribution of $p^{subset}_{k}$ will deviate more from $p^{dataset}_{k}$ than in the case of \textit{random sampling}.
However, here, on average, the number of objects will be higher.
We will evaluate this assumption in Section \ref{ssec:monte_carlo}.


The last algorithm is a~version of random sampling per class, which directly maximises the number of objects of a~given class -- \textit{maximum object number sampling per class} -- MONSPeC.
The steps of the algorithm are as follows:
\begin{enumerate}
    \item For each class, create a~set $idx_k$ which includes the indexes of training samples containing at least one object of that class.
    \item From each set $idx_k$, select $\frac{N}{C}$ samples that have the largest number of objects of class $k$.
\end{enumerate}
Similarly to \textit{random sampling per class}, duplicate samples may be present in the resulting subset.
Intuitively, the distribution of $p^{subset}_{k}$ for this algorithm will deviate most from $p^{dataset}_{k}$.
However, the number of objects in this case will be maximal.
These assumptions, as for random sampling per class, will be verified in Section \ref{ssec:monte_carlo}.

We decided to compare these three algorithms because of their efficiency and speed.
There are several reasons why we did not consider more complex approaches.
Firstly, to the best of our knowledge, there is no work that focuses on subset selection for the task of object detection based on point clouds.
The sole reference available involves an approach from OpenPCDet \cite{open_pc_det} in which models are trained on a~completely random Waymo subset, representing 20\% of the total dataset size.
This most basic subset selection algorithm is included in our analysis.
Among image-based algorithms, there are also few approaches that focus on the task of object detection.
They usually employ ensemble models, incorporating a sample into the dataset when there is a significant difference in the models' results. 
However, these ensemble-based approaches, e.g. \cite{al_det_ensemble}, require multiple model trainings, leading to time-consuming subset generation without necessarily improving efficiency.
We initially tested the \textit{build-up} approach from \cite{al_det_ensemble}, but it ended up giving slightly worse results than \textit{random sampling per-class} with incomparably more computational and time effort.
With a~subset size of 20\% nuScenes, the \textit{random sampling per class} took less than a~millisecond, while the approach with \cite{al_det_ensemble} took over 40h on a~12-core AMD Ryzen 9 3900X processor with two Nvidia RTX 3080 GPUs.
In contrast, the other approaches are adapted to the classification task \cite{nas_cls_proxy1} or inaccurately described as \cite{nas_det_fcos}.


In Section \ref{sec:experiments} we will evaluate each method -- select subsets of several fixed sizes with each method, train the PointPillars and CenterPoint-Pillar networks on each, and compare the resulting detection efficiencies.
In addition, we will check with the Monte Carlo method the distribution of the distance $p^{subset}_{k}$ from $p^{dataset}_{k}$ and the distribution of the number of objects of each class.

\section{Experiments}
\label{sec:experiments}


The first stage of the experiments is to estimate for each subset size the distribution of the L1 distance between $p^{subset}_k$ and $p^{dataset}_k$ and the distribution of the number of objects of each class.
In addition, we measure average sampling time for each considered method.
In the second stage, we will select, using each of the three algorithms (described in the section \ref{ssec:algorithms_choosing_subset}), subsets of each of the fixed counts, train the PointPillars and CenterPoint-Pillar networks on each of them, and compare the obtained detection efficiencies.

\subsection{Estimating subset selection algorithms characteristics using Monte Carlo methods}
\label{ssec:monte_carlo}

The Monte Carlo method is a well-known technique for the mathematical modelling of complex processes.
We will apply it to estimate the distribution of the number of objects of each class and the distribution of the L1 distance between $p^{subset}_k$ and $p^{dataset}_k$.
We will determine the number of objects normalised by the expected number of objects returned by \textit{random sampling}, i.e.
$n^{norm}_k = \frac{n_k}{\bar{n_k}} = \frac{n_k}{\frac{N}{D} \times d_k}$.
Thus, we make the value of the number of objects independent of the size of the subset.
To further facilitate data analysis, values $n^{norm}_k$ for all classes we reduce to two values: $n^{norm}_{min} = \min_{\forall k = 1, ..., C} n^{norm}_{k}$ and $n^{norm}_{avg} = \frac{1}{C} \sum_ {k=1}^{k=C} n^{norm}_{k}$.
The value $n^{norm}_{min}$ informs whether there is enough learning data for each class.
On the other hand, the value $n^{norm}_{avg}$ informs about the average relative number of objects.
By representing the number of objects with these two indicators, it will be easy to compare the results between them for different subset sizes.
On the other hand, the distance L1 in our case is given by the formula: $L1 = \sum_ {k=1}^{k=C} |p^{dataset}_{k} - p^{subset}_k|$.


In this experiment, we consider subsets representing 5\%, 10\%, 20\%, 40\% and 80\% of the nuScenes training set, respectively.
For the estimation, we will draw 1'000'000 subsets for very subset size using the algorithms \textit{random sampling} and \textit{random sampling per class}.
The MONSPeC algorithm is deterministic, so the values of $n^{norm}_{min}$, $n^{norm}_{avg}$ and $L1$ only need to be computed once for each subset size.


\begin{figure}[!t]
\centerline{\includegraphics[width=0.4\textwidth]{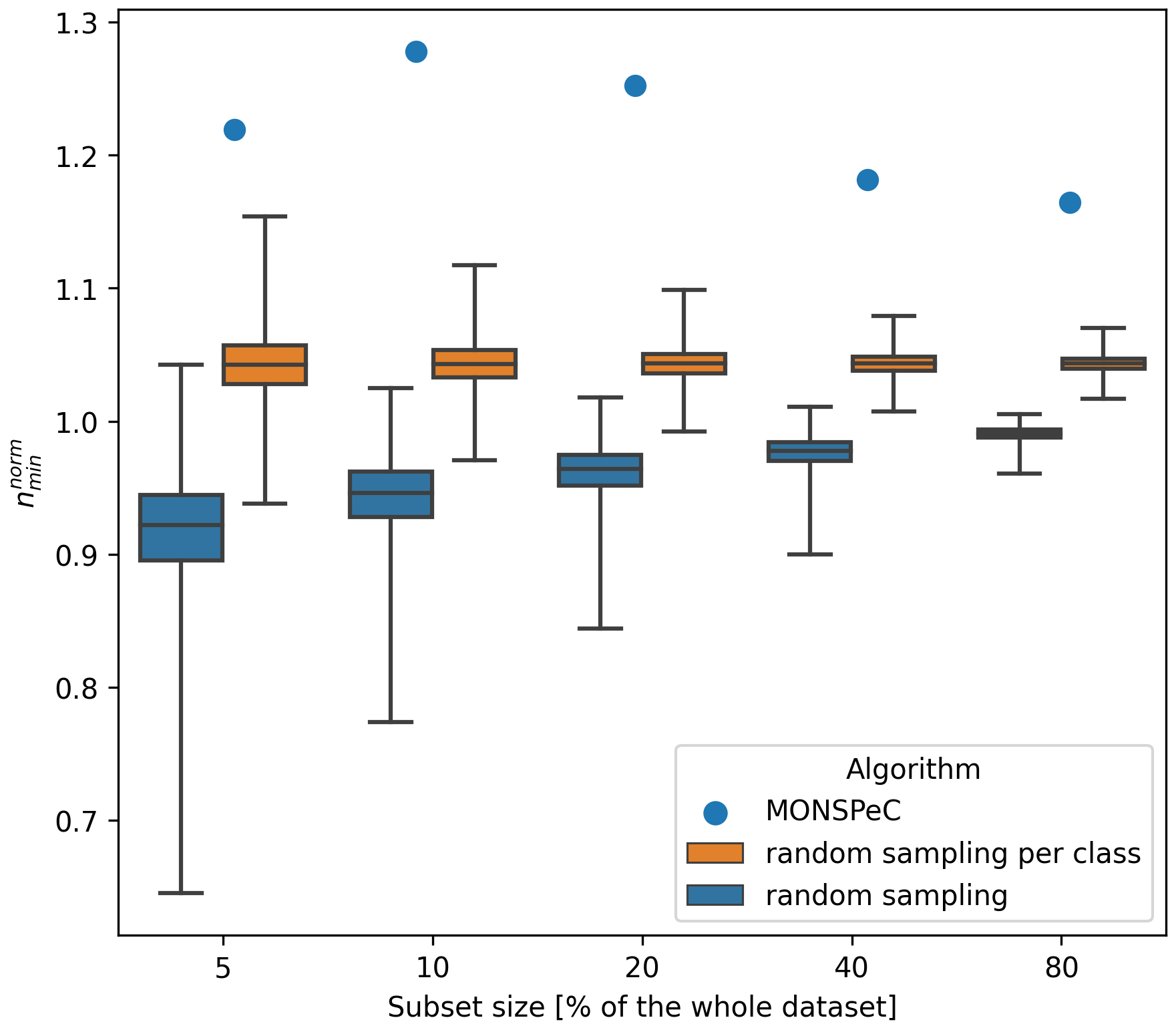}}
\caption{Box plot of $n^{norm}_{min}$ versus subset size for \textit{random sampling}, \textit{random sampling per class} and MONSPeC.
The distribution of the $n^{norm}_{min}$ values was estimated using the Monte Carlo method with 1'000'000 samples.}
\label{fig:mc_n_norm_min}
\end{figure}

\begin{figure}[!t]
\centerline{\includegraphics[width=0.4\textwidth]{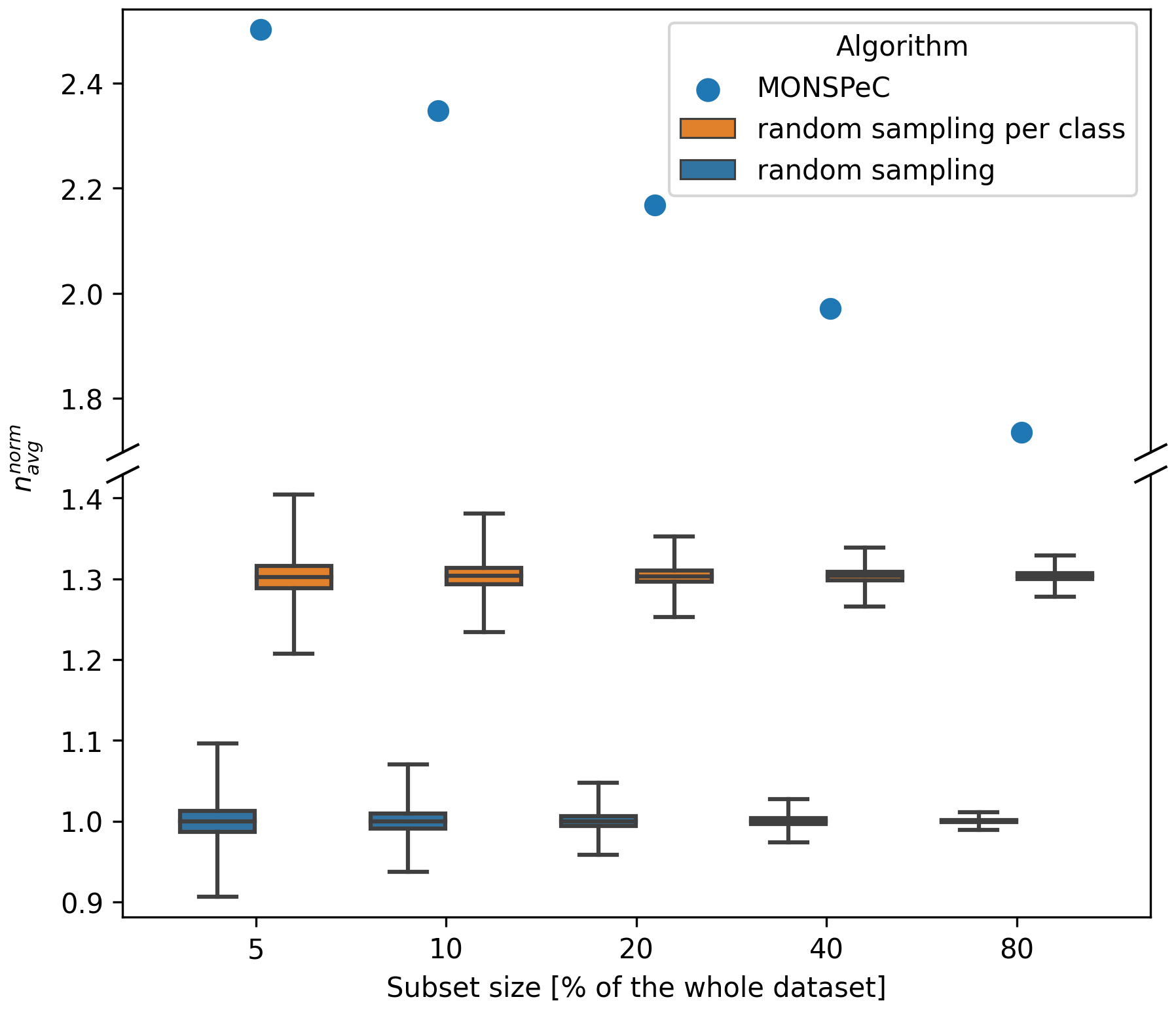}}
\caption{Box plot of $n^{norm}_{avg}$ versus subset size for \textit{random sampling}, \textit{random sampling per class} and MONSPeC.
The distribution of the $n^{norm}_{avg}$ values was estimated using the Monte Carlo method with 1'000'000 samples.}
\label{fig:mc_n_norm_avg}
\end{figure}

\begin{figure}[!t]
\centerline{\includegraphics[width=0.4\textwidth]{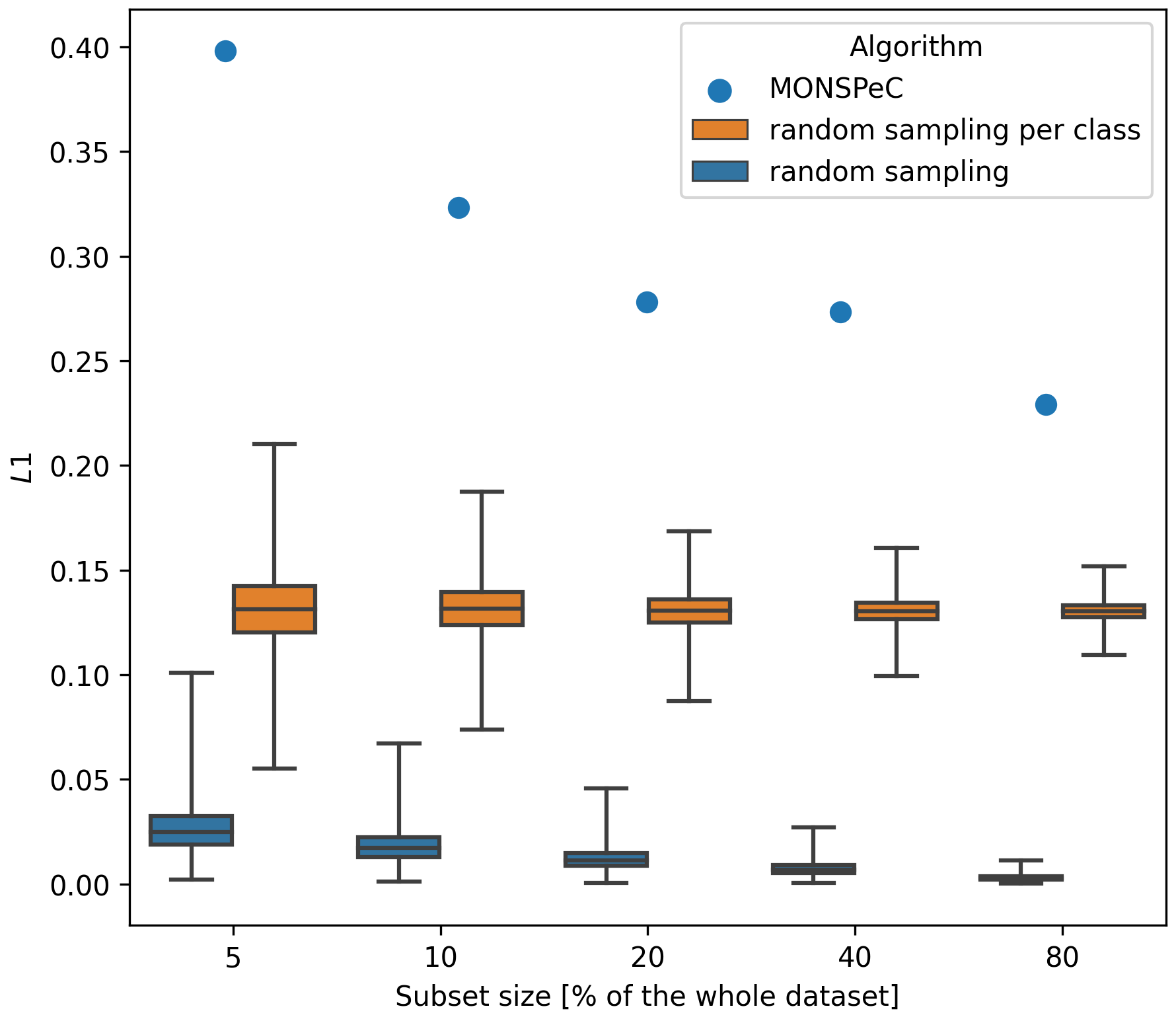}}
\caption{Box plot of the distance $L1$ between $p^{subset}_k$ and $p^{dataset}_k$ as a~function of subset size for \textit{random sampling}, \textit{random sampling per class} and MONSPeC.
The distribution of the $L1$ values was ascertained using the Monte Carlo method with 1'000'000 samples.}
\label{fig:mc_l1}
\end{figure}

Figure \ref{fig:mc_n_norm_min}, \ref{fig:mc_n_norm_avg}, and \ref{fig:mc_l1} show the distribution of $n^{norm}_{min}$, $n^{norm}_{avg}$, and $L1$ for each subset selection method, respectively.
For \textit{random sampling} and \textit{random per class sampling}, the results are presented as a~box-plot with whiskers.
The bottom and top edges of the box are the first and third quartiles, respectively, the line in the middle of the box is the median, and the whiskers cover the entire range of values -- from minimum to maximum.

The results presented here confirm the assumptions made in the Section \ref{ssec:algorithms_choosing_subset}:
\begin{itemize}
    \item The distance $p^{subset}_k$ from $p^{dataset}_k$ is statistically smallest for \textit{random sampling}, slightly larger for \textit{random per class sampling} and largest for MONSPeC.
    \item The average normalised number of objects $n^{norm}_{avg}$ is very large for MONSPeC -- the plot in Figure \ref{fig:mc_n_norm_avg} has the $y$ axis split into two ranges, as the values for MONSPeC differ significantly from the other two algorithms.
    Following MONSPeC, algorithms \textit{random sampling per class} and \textit{random sampling} rank second and third, respectively.
\end{itemize}
In addition, in Figure \ref{fig:mc_n_norm_min} all the boxes of the \textit{random sampling} method are noticeably below the line $n^{norm}_{min} = 1$.
It is therefore very likely that in the subset drawn by \textit{random sampling}, at least for one class the condition $n^{norm}_{k} < 1$ will be satisfied.
It has been verified experimentally that this probability is about 99.7\%.
This means that at least one class will have relatively little learning data.
If such a~situation occurs for a~rare class, then the already small number of objects of this class will be further reduced.
This, in turn, is likely to translate into poor detection performance for that class. 
Such a~situation is not likely to occur for the \textit{random sampling per class} and MONSPeC methods -- their boxes are above the line $n^{norm}_{min} = 1$.
Therefore, one can expect that the detection efficiency on subsets selected by the \textit{random sampling} method will be the lowest.

Sampling time for all methods was measured on 1000 runs on nuScenes dataset for different subset sizes.
On average, \textit{random sampling} takes 0.42 ms, \textit{random sampling per-class} 0.65 ms and \textit{MONSPeC} takes 11.26 ms.
The sampling needs to be done only once -- in comparison to training that takes several hours, sampling time of each method is insignificant.

\subsection{Validating subset sampling algorithms by training object detectors on selected subsets}
\label{ssec:evaluation}

In this experiment, we consider subsets representing 5\%, 10\% and 20\% of the nuScenes training set, respectively.
We use the \textit{mmdetection3d} \cite{mmdet3d} framework, slightly modified for our needs, to train PointPillars and CenterPoint-Pillar networks on the selected subsets.
The computer used to perform the training process includes a~12-core AMD Ryzen 9 3900X processor and two Nvidia RTX 3080 GPUs.

We train PointPillars for 24 epochs with batch size 1 per GPU and mixed-precision with loss scale of 512. 
We use AdamW optimiser with weight decay 0.01 and exponentially decaying learning rate with 1000 iterations of warm-up (0.001 rate), initial learning rate of 0.01 and 0.1x decay after 20th and 23rd epoch.
Meanwhile, CenterPoint-Pillar is trained for 20 epochs with batch size 4 per GPU.
We use AdamW optimiser with weight decay 0.01 and cosine annealing learning rate schedule with initial learning rate $10^{-3}$, rising to $10^{-2}$ after 8 epochs and then falling to $10^{-7}$ at the final 20th epoch.

\begin{figure}[!t]
\centerline{\includegraphics[width=0.4\textwidth]{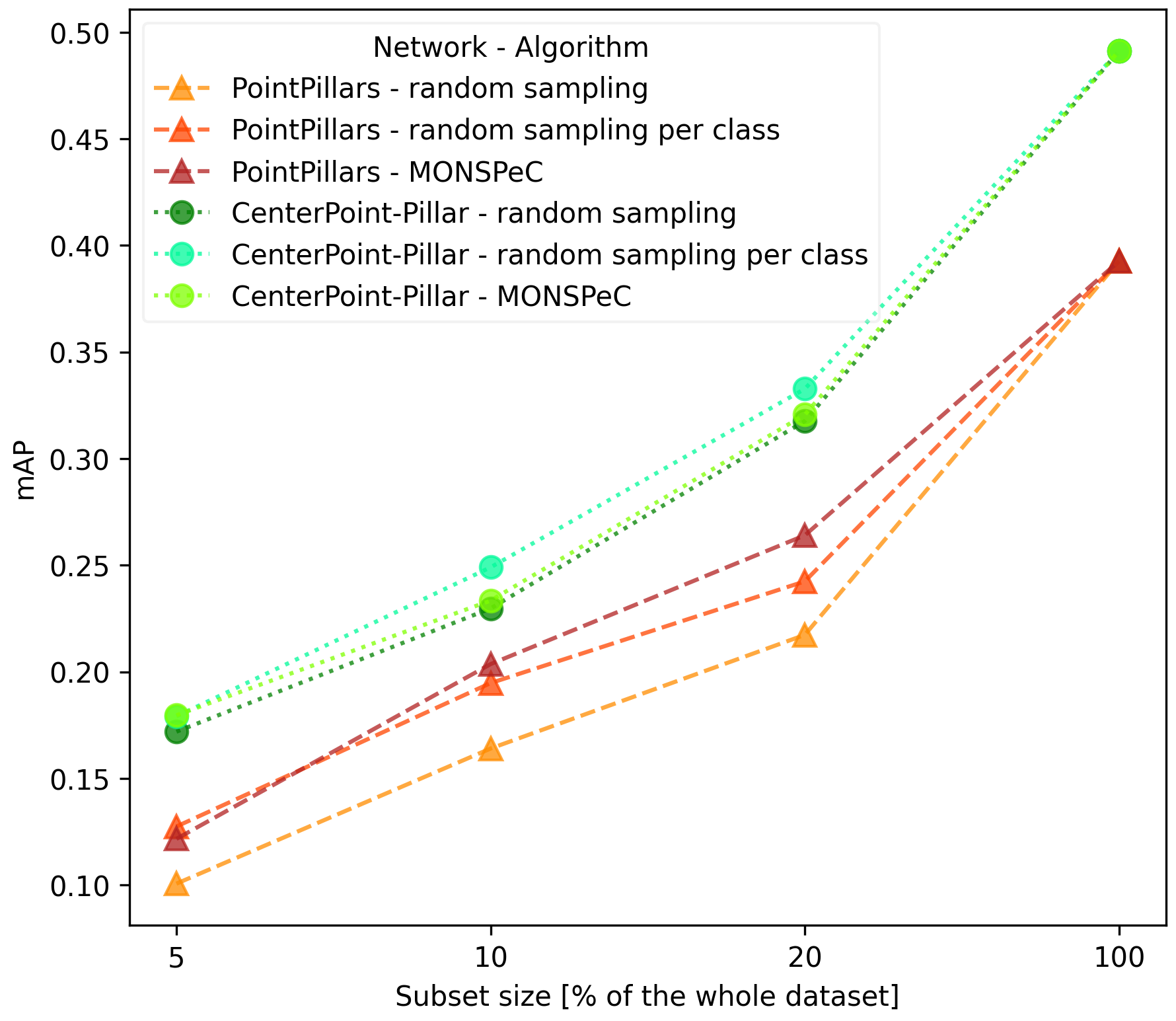}}
\caption{Plot of mAP against subset size, detection algorithm, and subset selection method.}
\label{fig:eval_map}
\end{figure}

\begin{figure}[!t]
\centerline{\includegraphics[width=0.4\textwidth]{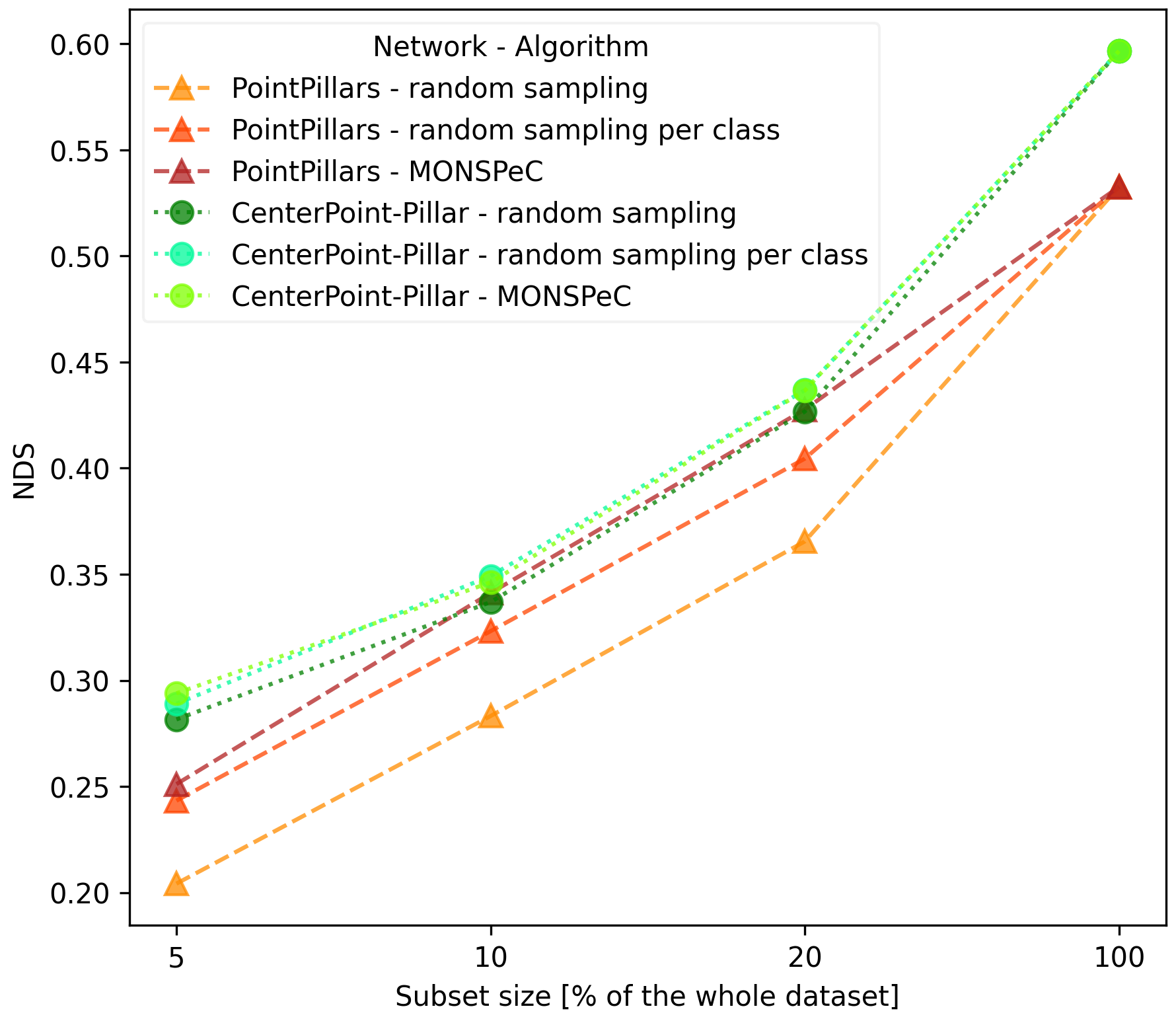}}
\caption{Plot of NDS against subset size, detection algorithm, and subset selection method.}
\label{fig:eval_nds}
\end{figure}

Figures \ref{fig:eval_map} and \ref{fig:eval_nds} illustrate the evaluation results of individual networks on subsets selected by the three considered algorithms.
Figure \ref{fig:eval_map} presents the mAP metric, while Figure \ref{fig:eval_nds} presents NDS\cite{nuscenes}.
The results for the whole dataset are included and marked as 100\% subset size.

In order for the method to be deemed effective, the condition 2 from Section \ref{sssec:objective} should be fulfilled, at least in the majority of the analysed scenarios.
For each sampling method, for each subset size, CenterPoint-Pillar is better than PointPillars, as it is for the whole dataset.
Therefore, condition 2 is met for all three methods.

As can be seen from the figures, the evaluation results on the subsets selected by the \textit{random sampling} method give the worst results -- both in terms of mAP and NDS.
In contrast, the results obtained with the \textit{random sampling per class} and MONSPeC methods are comparable.
For PointPillars, MONSPeC performs slightly better in terms of mAP and NDS.
For CenterPoint-Pillar, in terms of mAP, \textit{random sampling per class} performs slightly better, while in terms of NDS, both methods give virtually the same results.

Hence, it follows that \textit{random sampling} is the least efficient way to select a~subset among those considered.
Differences in speed between all the methods are insignificant, as was shown in Section \ref{ssec:monte_carlo}.
At the same time, the other two methods give better detection efficiency.
On the other hand, the choice between \textit{random sampling per class} and \textit{MONSPeC} is not obvious.
Their results are very close, sometimes one of them slightly dominates.
However, we lean more towards the MONSPeC due to the deterministic nature of its results. 
In this way, we can eliminate the risk of the subset selected by the \textit{random sampling per class} significantly deviating from the mean and returning exceptionally low detection accuracy.

\section{Conclusions}
\label{sec:conclusions}

We presented a~comparison of three algorithms for choosing a subset from the training set. 
We evaluated their effectiveness on the NuScenes dataset in terms of statistics and the detection performance achieved by two detection algorithms: PointPillars and CenterPoint-Pillar. 
We estimated their statistical properties using the Monte Carlo method. 

The obtained results indicate that a~completely random subset selection can be replaced by insignificantly slower and more efficient methods -- \textit{random sampling per class} or \textit{MONSPeC}. 
Out of these two algorithms, \textit{MONSPeC} is generally preferred due to its determinism.

Efficient selection of a~training set subset allows for faster prototyping of 3D object detection algorithms by reducing computation time on the GPU, thereby resulting in reduced energy consumption and associated environmental costs.

In our future work, we primarily plan to accelerate the prototyping of 3D object detection algorithms using either random sampling per class or MONSPeC. 
Additionally, we aim to evaluate the efficacy of the aforementioned subset selection algorithms on the KITTI and Waymo Open Dataset with a wider range of LiDAR object detection DCNN architectures, e.g. SECOND, PointRCNN or PV-RCNN.
Finally, we aim to explore alternative subset selection methods that consider attributes beyond the number of objects in each class, such as distribution of object sizes or orientations.

\bibliographystyle{IEEEtran}
\bibliography{references}

\end{document}